\documentclass[fleqn,10pt]{wlscirep}
\usepackage[utf8]{inputenc}
\usepackage[T1]{fontenc}
\usepackage{graphicx}
\usepackage{comment}
\usepackage{textcomp}
\usepackage{xcolor}
\usepackage{subcaption}
\usepackage{multirow}
\usepackage{bm}  
\title{P$^2$M: A \underline{P}rocessing-in-\underline{P}ixel-in-\underline{M}emory Paradigm for Resource-Constrained TinyML Applications}
\begin{document}

\author[1,*,+]{Gourav Datta}
\author[1,*]{Souvik Kundu}
\author[1,*]{Zihan Yin}
\author[1]{Ravi Teja Lakkireddy}
\author[2]{Joe Mathai}
\author[2]{Ajey P. Jacob}
\author[1]{Peter A. Beerel}
\author[1,2]{Akhilesh R. Jaiswal}
\affil[1]{Ming Hsieh Department of Electrical and Computer Engineering, University of Southern California, USA}
\affil[2]{Information Sciences Institute, University of Southern California, USA}
\affil[*]{these authors contributed equally to this work}
\affil[+]{Corresponding author: gdatta@usc.edu}


\begin{abstract}

The demand to process vast amounts of data generated from state-of-the-art high resolution cameras has motivated novel energy-efficient on-device AI solutions. Visual data in such cameras are usually captured in analog voltages by a sensor pixel array, and then converted to the digital domain for subsequent AI processing using analog-to-digital converters (ADC). Recent research has tried to take advantage of massively parallel low-power analog/digital computing in the form of near- and in-sensor processing, in which the AI computation is performed partly in the periphery of the pixel array and partly in a separate on-board CPU/accelerator. Unfortunately, high-resolution input images still need to be streamed  between the camera and the AI processing unit, frame by frame, causing energy, bandwidth, and security bottlenecks. To mitigate this problem, we propose a novel Processing-in-Pixel-in-memory (P$^2$M) paradigm, that customizes the pixel array by adding support for analog multi-channel, multi-bit convolution, batch normalization, and ReLU (Rectified Linear Units). Our solution includes a holistic algorithm-circuit co-design approach and the resulting P$^2$M paradigm can be used as a drop-in replacement for embedding memory-intensive first few layers of convolutional neural network (CNN) models within foundry-manufacturable CMOS image sensor platforms. Our experimental results indicate that P$^2$M reduces data transfer bandwidth from sensors and analog to digital conversions by ${\sim}21\times$, and the energy-delay product (EDP) incurred in processing a MobileNetV2 model on a TinyML use case for visual wake words dataset (VWW) by up to $\mathord{\sim}11\times$ compared to standard near-processing or in-sensor implementations, without any significant drop in test accuracy. 

\end{abstract}

\flushbottom
\maketitle
%
%
\thispagestyle{empty}

\section{Introduction}

Today's widespread applications of computer vision spanning surveillance \cite{surveillance}, disaster management \cite{disaster_management}, camera traps for wildlife monitoring \cite{camera_traps},  autonomous driving, smartphones, etc., are fueled by the remarkable technological advances in image sensing platforms \cite{sensor_increasing} and the ever-improving field of deep learning algorithms \cite{DL1}.
However, hardware implementations of vision sensing and vision processing platforms have traditionally been physically segregated. For example, current vision sensor platforms based on CMOS technology act as transduction entities that convert incident light intensities into digitized pixel values, through a two-dimensional array of photodiodes \cite{image_sensor}. The vision data generated from such CMOS Image Sensors (CIS) are often processed elsewhere in a cloud environment consisting of CPUs and GPUs \cite{Buckler2017ReconfiguringT}. The physical segregation of vision sensing and computing platforms leads to multiple bottlenecks concerning throughput, bandwidth, and energy-efficiency. 

To address these bottlenecks, many researchers are trying to bring intelligent data processing closer to the source of the vision data, \textit{i.e.,} closer to the CIS, taking one of three broad approaches -
near-sensor processing \cite{pinkhan2021jetcas,sony2020vision}, in-sensor processing \cite{chen2020pns}, and in-pixel processing \cite{Mennel2020UltrafastMV,scamp2020eccv,song2021reconfigurable}. Near-sensor processing aims to incorporate a dedicated machine learning accelerator chip on the same printed circuit board \cite{pinkhan2021jetcas}, or even 3D-stacked with the CIS chip \cite{sony2020vision}. Although this enables processing of the CIS data closer to the sensor rather than in the cloud, it still suffers from 
the data transfer costs between the CIS and processing chip. On the other hand, in-sensor processing solutions \cite{chen2020pns} integrate digital or analog circuits within the periphery of the CIS sensor chip, reducing the data transfer between the CIS sensor and processing chips. Nevertheless, these approaches still often require data to be streamed (or read in parallel) through a bus from CIS photo-diode arrays into the peripheral processing circuits \cite{chen2020pns}. In contrast, in-pixel processing solutions, such as \cite{jaiswal1, jaiswal2, Mennel2020UltrafastMV,scamp2020eccv,song2021reconfigurable}, aim to embed processing capabilities within the individual CIS pixels. 
Initial efforts have focused on in-pixel analog convolution operation \cite{jaiswal1, jaiswal2} but many \cite{jaiswal1, jaiswal2, Mennel2020UltrafastMV,angizi2022pisa} require the use of emerging non-volatile memories or 2D materials.
Unfortunately, these technologies are not yet mature and thus not amenable to the existing foundry-manufacturing of CIS. Moreover, these works fail to support multi-bit, multi-channel convolution 
operations, batch normalization (BN), and Rectified Linear Units (ReLU) needed for most practical deep learning applications.
Furthermore, works that target digital CMOS-based in-pixel hardware, organized as pixel-parallel single instruction multiple data (SIMD) processor arrays \cite{scamp2020eccv}, do not support convolution operation, and are thus limited to toy workloads, such as digit recognition.
Many of these works rely on digital processing which typically yields lower levels of parallelism compared to their analog in-pixel alternatives. In contrast, the work in \cite{song2021reconfigurable}, leverages in-pixel parallel analog computing, wherein the weights of a neural network are represented as the exposure time of individual pixels. 
Their approach requires weights to be made available for manipulating pixel-exposure time through control pulses, leading to a data transfer bottleneck between the weight memories and the sensor array. Thus, 
an in-situ CIS processing solution where both the weights and input activations are available within individual pixels that efficiently implements critical deep learning operations such as multi-bit, multi-channel convolution, BN, and ReLU operations has 
remained elusive. Furthermore, all existing in-pixel computing solutions have targeted datasets that do not represent realistic applications of machine intelligence mapped onto state-of-the-art CIS. Specifically, most of the existing works are focused on simplistic datasets like MNIST \cite{scamp2020eccv}, while few \cite{song2021reconfigurable} use the CIFAR-$10$ dataset which has input images with a significantly low resolution ($32{\times}32$), that does not represent images captured by state-of-the-art high resolution CIS.

Towards that end, we propose a novel in-situ computing paradigm at the sensor nodes called \textit{Processing-in-Pixel-in-Memory} (P$^2$M), that incorporates both the network weights and activations to enable massively parallel, high-throughput intelligent computing inside CISs. 
In particular, our circuit architecture not only enables in-situ multi-bit, multi-channel, dot product analog acceleration needed for convolution, but re-purposes the on-chip digital \textit{correlated double sampling} (CDS) circuit and single slope ADC (SS-ADC) typically available in conventional CIS to implement \textit{all the required computational aspects for the first few layers} of a state-of-the-art deep learning network. Furthermore, the proposed architecture is coupled with a circuit-algorithm co-design paradigm that captures the circuit non-linearities, limitations, and bandwidth reduction goals for improved latency and energy-efficiency. The resulting paradigm is the first to demonstrate feasibility for enabling complex, intelligent image processing applications (beyond toy datasets), on high resolution images of Visual Wake Words (VWW) dataset, catering to a real-life TinyML application. We choose to evaluate the efficacy of P$^2$M on TinyML applications, as they impose tight compute and memory budgets, that are otherwise difficult to meet with current in- and near-sensor processing solutions, particularly for high-resolution input images. 
Key highlights of the presented work are as follows:

\begin{figure}[!t]
\centering
\includegraphics[width = 0.8\linewidth]{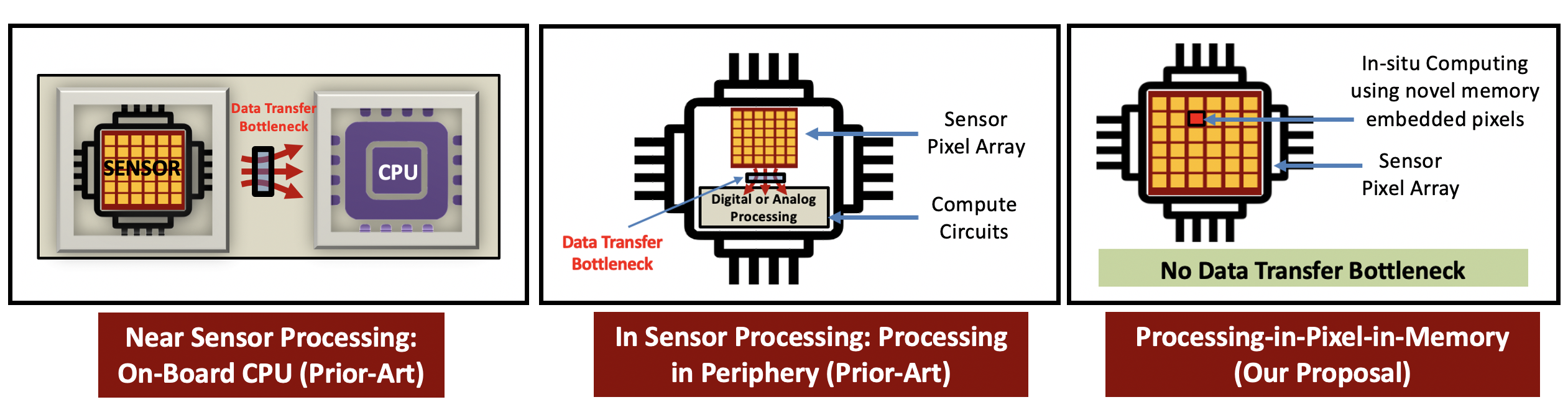}
\caption{Existing and Proposed Solutions to alleviate the energy, throughput, and bandwidth bottleneck caused by the segregation of \textit{Sensing} and \textit{Compute}.}
\label{fig:pip_initials}
\vspace{-1mm}
\end{figure}

\begin{enumerate}
    \item We propose a novel processing-in-pixel-in-memory (P$^2$M) paradigm for resource-constrained sensor intelligence applications, wherein novel memory-embedded pixels enable massively parallel dot product acceleration using in-situ input activations (photodiode currents) and in-situ weights all available within individual pixels.
    \item We propose re-purposing of on-chip memory-embedded pixels, CDS circuits and SS-ADCs to implement positive and negative weights, BN, and digital ReLU functionality within the CIS chip, thereby mapping all the computational aspects for the first few layers of a complex state-of-the-art deep learning network within CIS.
    \item We further develop a compact MobileNet-V$2$ based model optimized specifically for P$^2$M-implemented hardware constraints, and benchmark its accuracy and energy-delay product (EDP) on the VWW dataset, which represents a common use case of visual TinyML.
\end{enumerate}

The remainder of the paper is organized as follows. Section \ref{sec:opp_challenges} discusses the challenges and opportunities for 
P$^2$M. Section \ref{sec:circuit_architecture} explains our proposed P$^2$M circuit implementation using manufacturable memory technologies. Then, Section \ref{sec:algo_HW_codesign} discusses our approach for P$^2$M-constrained algorithm-circuit co-design. Section \ref{sec:results} presents our TinyML benchmarking dataset, model architectures, test accuracy and EDP results. Finally, some conclusions are provided in Section \ref{sec:conc}.

\section{Challenges \& Opportunities in P$^2$M}\label{sec:opp_challenges}

The ubiquitous presence of CIS-based vision sensors has driven the need to enable machine learning computations closer to the sensor nodes. However, given the computing complexity of modern CNNs, such as Resnet-18 \cite{he2015deep} and SqueezeNet \cite{iandola2016squeezenet}, it is not feasible to execute the entire deep-learning network, including all the layers within the CIS chip. As a result, recent intelligent vision sensors, for example, from Sony \cite{sony2020vision}, which is equipped with basic AI processing functionality (e.g., computing image metadata), features a multi-stacked configuration consisting of separate pixel and logic chips that must rely on high and relatively energy-expensive inter-chip communication bandwidth. 

Alternatively, we assert that embedding part of the deep learning network within pixel arrays in an in-situ manner can lead to a significant reduction in data bandwidth (and hence energy consumption) between sensor chip and downstream processing for the rest of the convolutional layers. This is because the first few layers of carefully designed CNNs, as explained in Section \ref{sec:algo_HW_codesign}, can have a significant compressing property, i.e., the output feature maps have reduced bandwidth/dimensionality compared to the input image frames. In particular, our proposed P$^2$M paradigm enables us to map all the computations of the first few layers of a CNN into the pixel array. The paradigm includes a holistic hardware-algorithm co-design framework that captures the specific circuit behavior, including circuit non-idealities, and hardware limitations, during the design, optimization, and training of the proposed machine learning networks. The trained weights for the first few network layers are then mapped to specific transistor sizes in the pixel-array. 
Because the transistor widths are fixed during manufacturing, the corresponding CNN weights lack programmability. Fortunately, it is common to use the pre-trained versions of the first few layers of modern CNNs as high-level feature extractors are common across many vision tasks \cite{cnn_features}.
Hence, the fixed weights in the first few CNN layers do not limit the use of our proposed scheme for a wide class of vision applications. 
Moreover, we would like to emphasize that the memory-embedded pixel also work seamlessly well by replacing fixed transistors with emerging non-volatile memories, as described in Section \ref{subsec:area}.
Finally, the presented P$^2$M paradigm can be used in conjunction with existing near-sensor processing approaches for added benefits, such as, improving the energy-efficiency of the remaining convolutional layers. 

\section{P$^2$M Circuit Implementation}\label{sec:circuit_architecture}

\begin{figure}[!t]
\centering
\includegraphics[width = 0.9\linewidth]{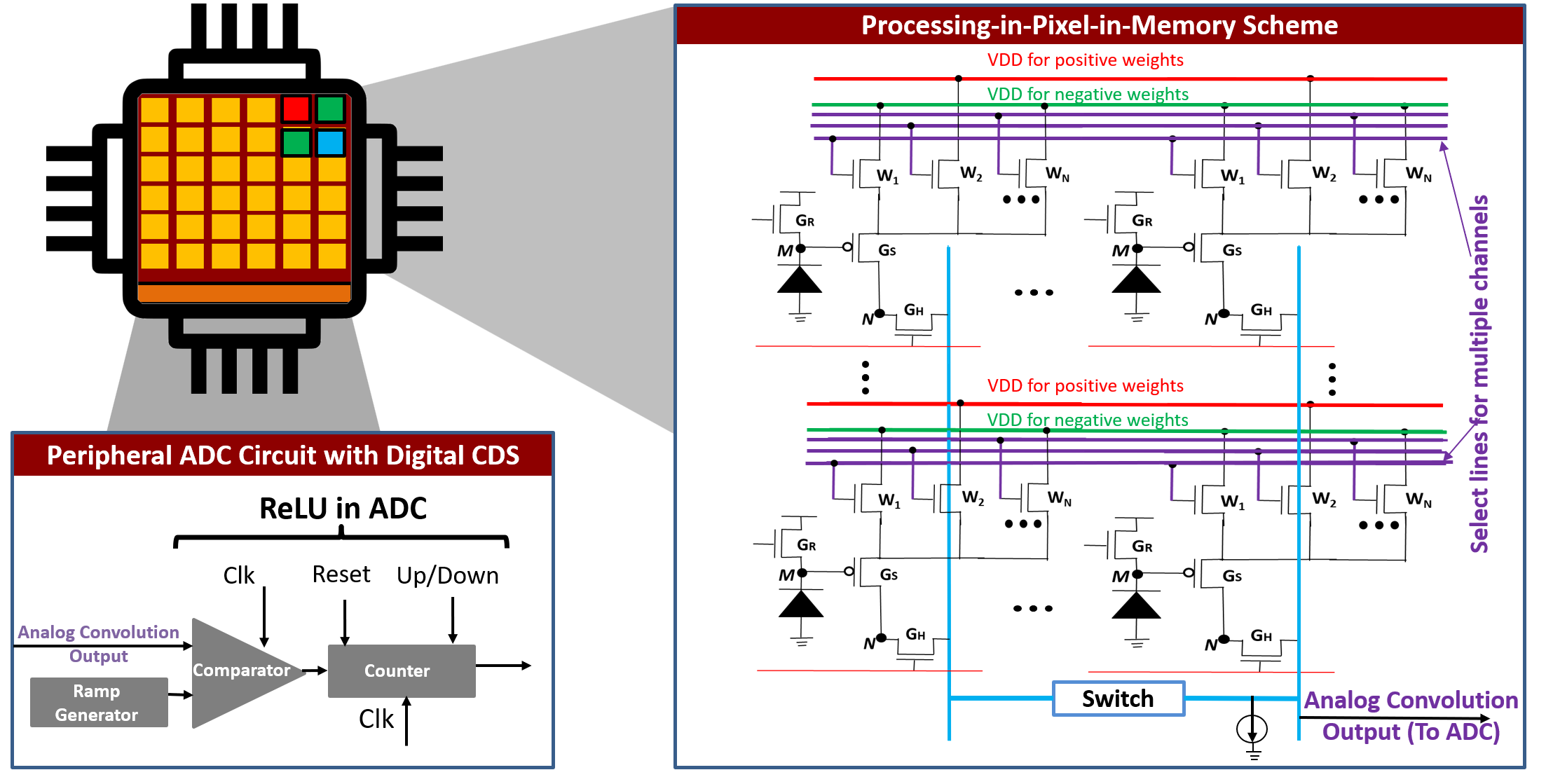}
\caption{Proposed circuit techniques based on presented P$^2$M scheme capable of mapping all computational aspects for the first few layers of a modern CNN layer within CIS pixel arrays.}
\label{fig:pip_circuit}
\vspace{-1mm}
\end{figure}

This section describes key circuit innovations that enable us to embed 
all the computational aspects for the first few layers of a complex CNN architecture within the CIS. 
An overview of our proposed pixel array that enables the availability of weights and activations within individual pixels with 
appropriate peripheral circuits is shown in Fig. \ref{fig:pip_circuit}. 

\subsection{Multi-Channel, Multi-Bit Weight Embedded Pixels}\label{subsec:embedded_memory}

Our modified pixel circuit builds upon the standard three transistor pixel by embedding additional transistors $W_i$s that represent weights of the CNN layer, as shown in Fig. \ref{fig:pip_circuit}. Each weight transistor $W_i$ is connected in series with the source-follower transistor $G_s$. When a particular weight transistor $W_i$ is activated (by pulling its gate voltage to $V_{DD}$), the pixel output is modulated both by the driving strength of the transistor $W_i$ and the voltage at the gate of the source-follower transistor $G_s$. Thus, the pixel output performs an approximate multiplication operation between the input light intensity (voltage at the gate of transistor $G_s$) and the weight (or driving strength) of the transistor $W_i$. Multiple weight transistors $W_i$s are incorporated within the same pixel and are controlled by independent gate control signals. These weight transistors can be used to
implement different channels in the output feature map of the layer. Thus, the gate signals represent select lines for specific channels in the output feature map.

The presented circuit can support both overlapping and non-overlapping strides depending on the number of weight transistors $W_i$s per pixel. Specifically, each stride for a particular kernel can be mapped to a different set of weight transistors over the pixels (input activations).
The transistors $W_i$s represent multi-bit weights as the driving strength of the transistors can be controlled over a wide range based on transistor width, length, and threshold voltage. 

\subsection{In-situ Multi-pixel Convolution Operation}

To achieve the convolution operation, we simultaneously activate multiple pixels. In the specific case of VWW, we activate $X{\times}Y{\times}3$ pixels at the same time, where $X$ and $Y$ denote the spatial dimensions and $3$ corresponds to the RGB (red, blue, green) channels in the input activation layer. For each activated pixels, the pixel output is modulated by the photo-diode current and the weight of the activated $W_i$ transistor associated with the pixel. The weight transistors are activated by respective select lines connected to their gates. The weight transistors $W_i$ represent multi-bit weight through its driving strength. Outputs from multiple pixel combine with each other, thereby, mimicking an accumulation operation. Thus, the voltage at the output of the select lines, shown as vertical blue lines in Fig. \ref{fig:pip_circuit}, represent the convolution operation between input activations and the stored weight inside the pixel. Note, in order to generate multiple output feature maps, the convolution operation has to be repeated for each channel in the output feature map. The corresponding weight for each channel is stored in a separate weight transistor embedded inside each pixel. Thus, there are as many weight transistors embedded within a pixel as there are number of channels in the output feature map. 

In summary, the presented scheme can perform in-situ multi-bit, multi-channel analog convolution operation inside the pixel array, wherein both input activations and network weights are present within individual pixels.

\begin{figure}[!b]
\centering
\begin{subfigure}{0.45\textwidth}
    \includegraphics[width = \textwidth]{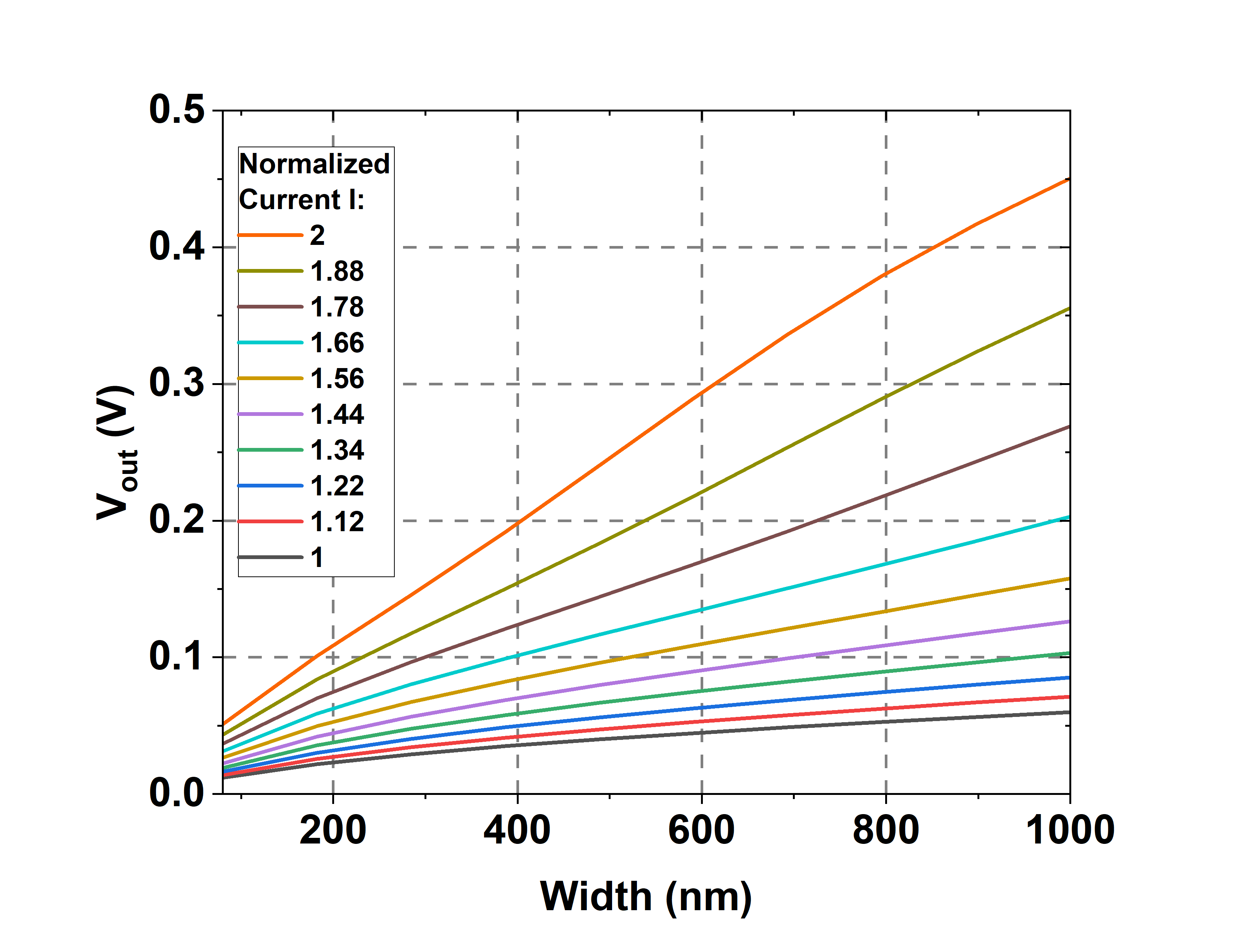}
    \caption{ }
    \label{fig:p2m_WI}
    
\end{subfigure}
\hfill
\begin{subfigure}{0.45\textwidth}
    \includegraphics[width = \textwidth]{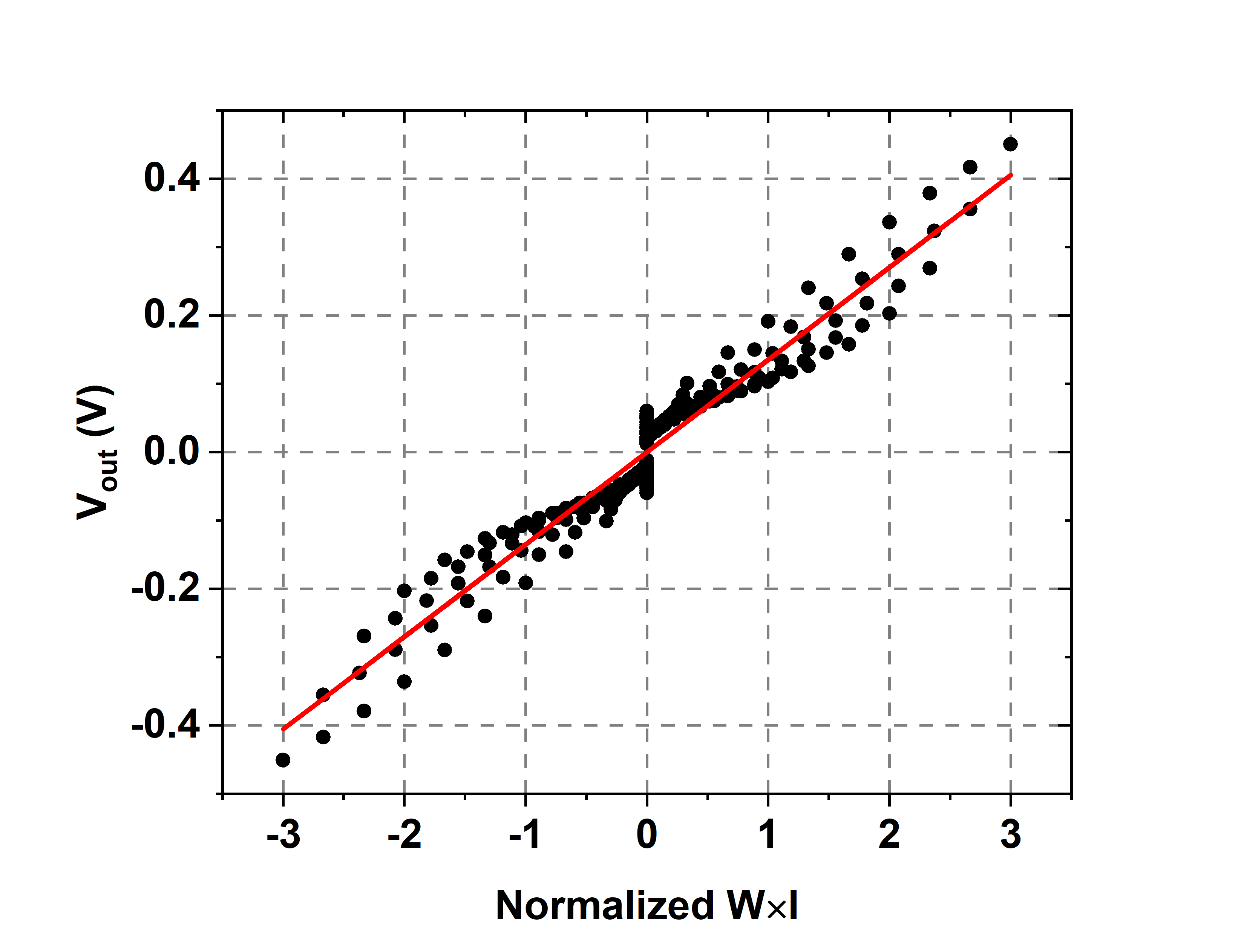}
    \caption{ }
    \label{fig:p2m_WI}
\end{subfigure}
\hfill
\caption{(a) Pixel output voltage as a function of weight (transistor width) and input activation (Normalized photo-diode current) simulated on Globalfoundries $22$nm FD-SOI node. As expected pixel output increases both as a function of weights and input activation. (b) A scatter plot comparing pixel output voltage to ideal multiplication value of Weights$\times$Input activation (Normalized $W\times I$). The plot confirms that the output pixel voltage from each pixel represents approximate product of Weights$\times$Input activation. }
\label{P2M_circuitall}
\vspace{-1mm}
\end{figure}

\subsection{Re-purposing Digital Correlated Double Sampling Circuit and Single-Slope ADCs as ReLU Neurons}\label{subsec:relu_circuit}

Weights in a CNN layer span positive and negative values. As discussed in the previous sub-section, weights are mapped by the driving strength (or width) of transistors $W_i$s. As the width of transistors cannot be negative, the $W_i$ transistors themselves cannot represent negative weights. Interestingly, we circumvent this issue by re-purposing on-chip digital CDS circuit present in many state-of-the-art commercial CIS \cite{cds1,cds2}. A digital CDS is usually implemented in conjunction to column parallel Single Slope ADCs (SS-ADCs). A single slope ADC consists of a ramp-generator, a comparator, and a counter (see Fig. \ref{fig:pip_circuit}). An input analog voltage is compared through the comparator to a ramping voltage with a fixed slope, generated by the ramp generator. A counter which is initially reset, and supplied with an appropriate clock, keeps counting until the ramp voltage crosses the analog input voltage. At this point, the output of counter is latched and represents the converted digital value for input analog voltage. A traditional CIS digital CDS circuit takes as input two correlated samples at two different time instances. The first sample corresponds to the reset noise of the pixel and the second sample to the actual signal superimposed with the reset noise. A digital CIS CDS circuit then takes the difference between the two samples, thereby, eliminating reset noise during ADC conversion. In an SS-ADC the difference is taken by simply making the counter `up' count for one sample and `down' count for the second.

\begin{figure}[!b]
\centering
\begin{subfigure}{0.45\textwidth}
    \includegraphics[width = \textwidth]{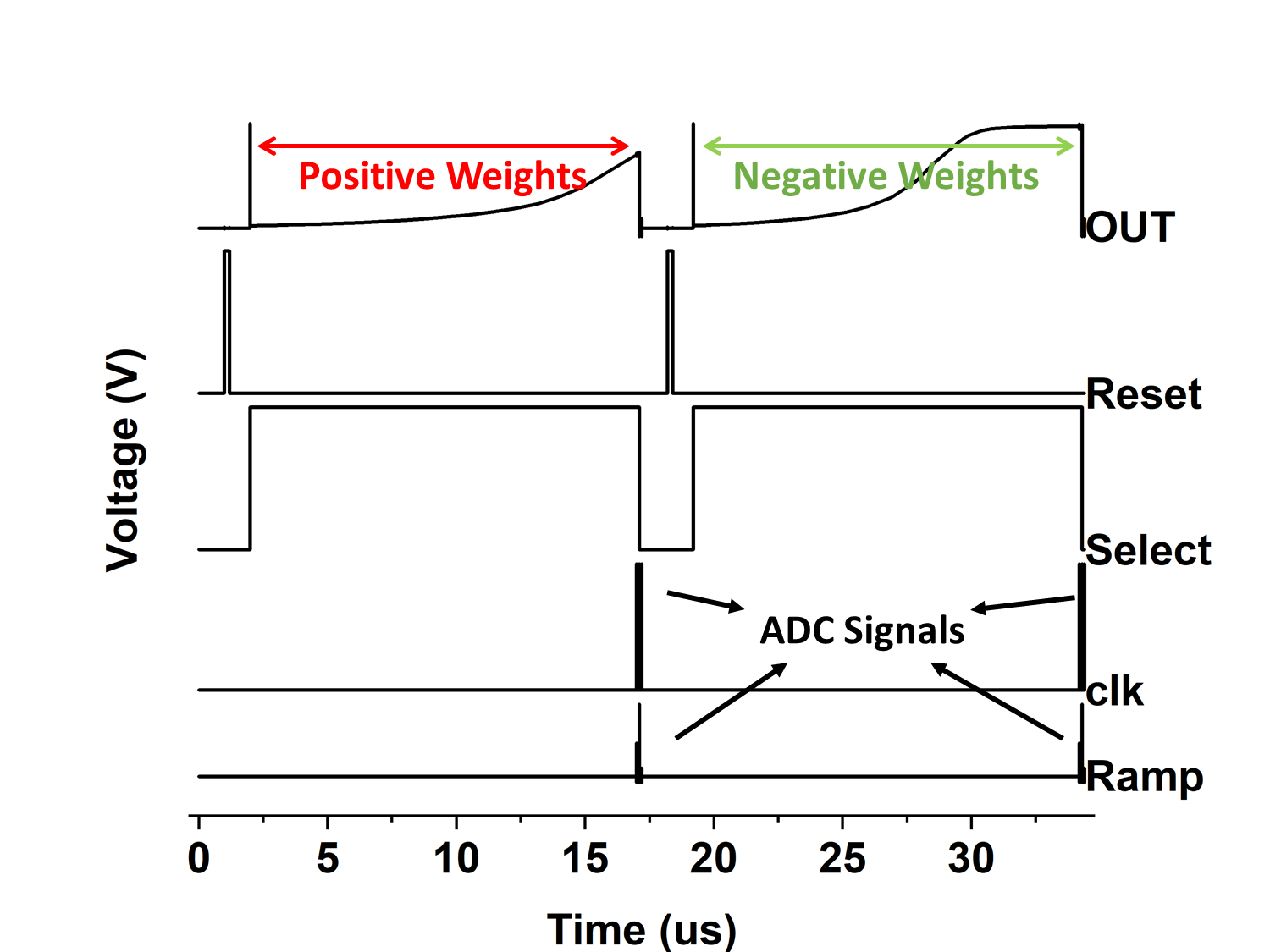}
    \caption{ }
    \label{fig:p2m_wave}
    
\end{subfigure}
\begin{subfigure}{0.45\textwidth}
    \includegraphics[width = \textwidth]{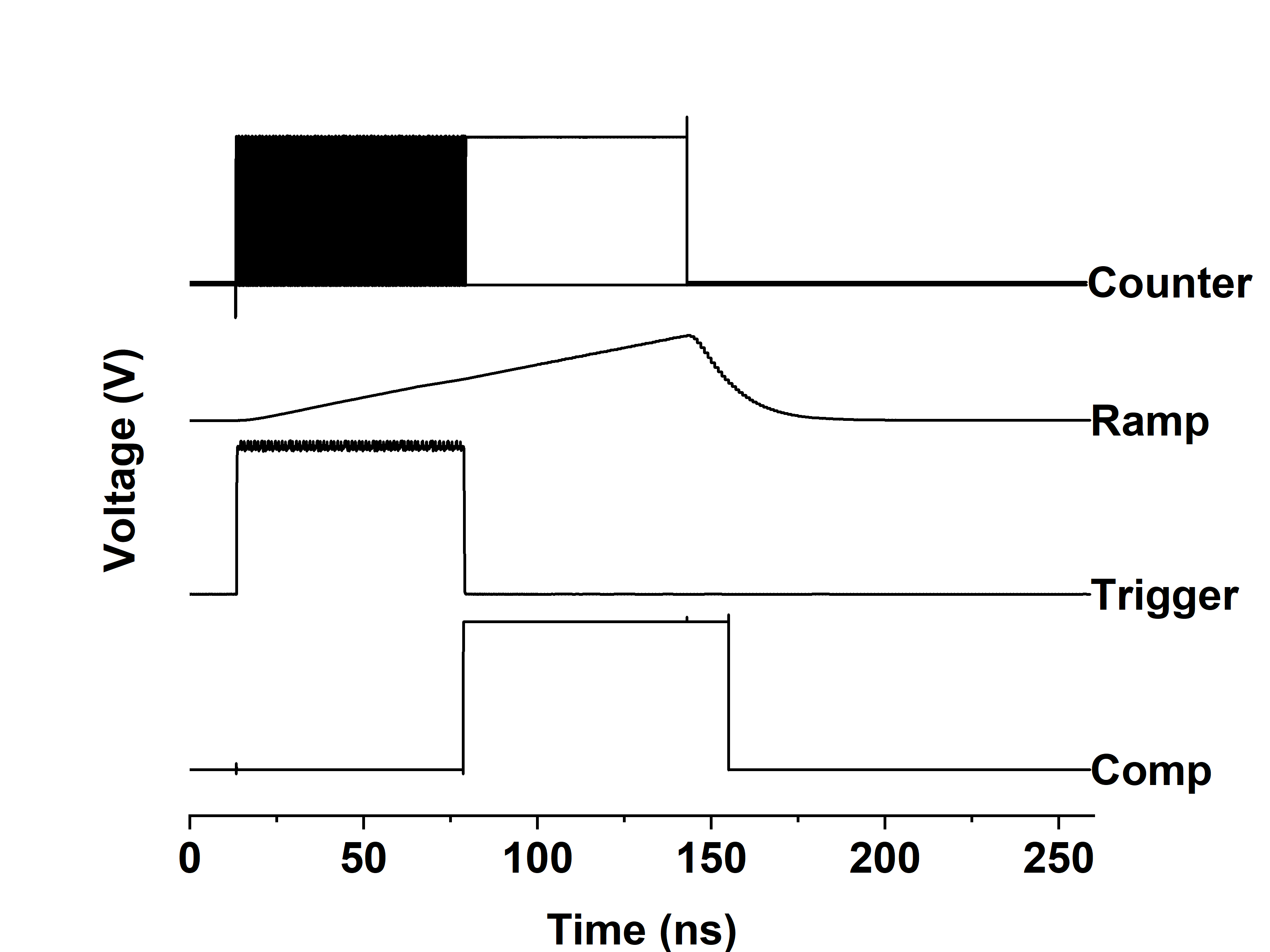}
    \caption{ }
    \label{fig:p2m_adc}
\end{subfigure}
\hfill
\caption{(a) A typical timing waveform, showing double sampling (one for positive and other for negative) weights. (b) Typical timing waveform for the SS-ADC showing comparator output (Comp), counter enable (trigger), ramp generator output, and counter clock (Counter). }
\label{P2M_circuitall}
\vspace{-1mm}
\end{figure}

We utilize the noise cancelling, differencing behavior of the CIS digital CDS circuit already available on commercial CIS chips to implement positive and negative weights and implement ReLU. First, each weight transistor embedded inside a pixel is `tagged' as a positive or a `negative weight' by connecting it to `red lines' (marked as VDD for positive weights in Fig. \ref{fig:pip_circuit}) and `green lines' (marked as VDD for negative weights in Fig. \ref{fig:pip_circuit}). For each channel, we activate multiple pixels to perform an inner-product and read out two samples. The first sample corresponds to a high VDD voltage applied on the `red lines' (marked as VDD for positive weights in Fig. \ref{fig:pip_circuit}) while the `green lines' (marked as VDD for negative weights in Fig. \ref{fig:pip_circuit}) are kept at ground. The accumulated multi-bit dot product result is digitized by the SS-ADC, while the counter is `up' counting. The second sample, on the other hand, corresponds to a high VDD voltage applied on the `green lines' (marked as VDD for negative weights in Fig. \ref{fig:pip_circuit}) while the `red lines' (marked as VDD for positive weights in Fig. \ref{fig:pip_circuit}) are kept at ground. The accumulated multi-bit dot product result is again digitized and also subtracted from the first sample by the SS-ADC, while the counter is `down' counting. Thus, the digital CDS circuit first accumulates the convolution output for all positive weights and then subtracts the convolution output for all negative weights for each channel, controlled by respective select lines for individual channels. Note, possible sneak currents flowing between weight transistors representing positive and negative weights can be obviated by integrating a diode in series with weight transistors or by simply splitting each weight transistor into two series connected transistors, where the channel select lines control one of the series connected transistor, while the other transistor is controlled by a select line representing positive/negative weights. 

Interestingly, re-purposing the on-chip CDS for implementing positive and negative weights also allows us to easily implement a quantized ReLU operation inside the SS-ADC. ReLU clips negative values to zero. This can be achieved by ensuring that the final count value latched from the counter (after the CDS operation consisting of `up' counting and then `down' counting') is either positive or zero. Interestingly, before performing the dot product operation, the counter can be reset to a non-zero value representing the scale factor of the BN layer as described in Section \ref{sec:algo_HW_codesign}. Thus, by embedding multi-pixel convolution operation and re-purposing on-chip CDS and SS-ADC circuit for implementing positive/negative weights, batch-normalization and ReLU operation, our proposed P$^2$M scheme can implement all the computational aspect for the first few layers of a complex CNN within the pixel array enabling massively parallel in-situ computations.

Putting these features together, our proposed P$^2$M circuit computes one channel at a time and has three phases of operation:
\begin{enumerate}
    \item Reset Phase: First, the voltage on the photodiode node $M$ (see Fig. \ref{fig:pip_circuit}) is pre-charged or reset by activating the reset transistor $G_r$. Note, since we aim at performing multi-pixel convolution, the set of pixels $X{\times}Y{\times}3$ are reset, simultaneosuly.
    \item Multi-pixel Convolution Phase: Next, we discharge the gate of the reset transistor $G_r$ which deactivates $G_r$. Subsequently, $X{\times}Y{\times}3$ pixels are activated by pulling the gate of respective $G_H$ transistors to VDD.  Within the activated set of pixels, a single weight transistor corresponding to a particular channel in the output feature map is activated, by pulling high its gate voltage through the select lines (labeled as select lines for multiple channels in Fig. 2). As the photodiode is sensitive to the incident light, photo-current is generated as light shines upon the diode (for a duration equal to exposure time), and voltage on the gate of $G_s$ is modulated in accordance to the photodiode current that is proportional to the intensity of incident light. The pixel output voltage is a function of the incident light (voltage on node $M$) and the driving strength of the activated weight transistor within each pixel. Pixel output from multiple pixels are accumulated on the column-lines and represent the multi-pixel analog convolution output. The SS-ADC in the periphery converts analog output to a digital value.
    Note, the entire operation is repeated twice, one for positive weights (`up' counting) and another for negative weights (`down counting').
    \item ReLU Operation: Finally, the output of the counter is latched and represents a quantized ReLU output. It is ensured that the latched output is either positive or zero, thereby mimicking the ReLU functionality within the SS-ADC.
\end{enumerate}

The entire P$^2$M circuit is simulated using commercial $22$nm Globalfoundries FD-SOI (fully depleted silicon-on-insulator) technology, the SS-ADCs are implemented using a using a bootstrap ramp generator and dynamic comparators. Assuming the counter output which represents the ReLU function is an $N$-bit integer, it needs $2^N$ cycles for a single conversion. The ADC is supplied with a 2GHz clock for the counter circuit. SPICE simulations exhibiting the multiplicative nature of weight transistor embedded pixels with respect to photodiode current is shown in Fig. 3 (a)-(b). A typical timing waveform showing pixel operation along with SS-ADC operation simulated on $22$nm Globalfoundries technology node is shown in Fig. \ref{fig:p2m_wave}.

\subsection{CIS Process Integration and Area Considerations}\label{subsec:area}

\begin{figure}[h]
\centering
\includegraphics[width = 0.7\linewidth]{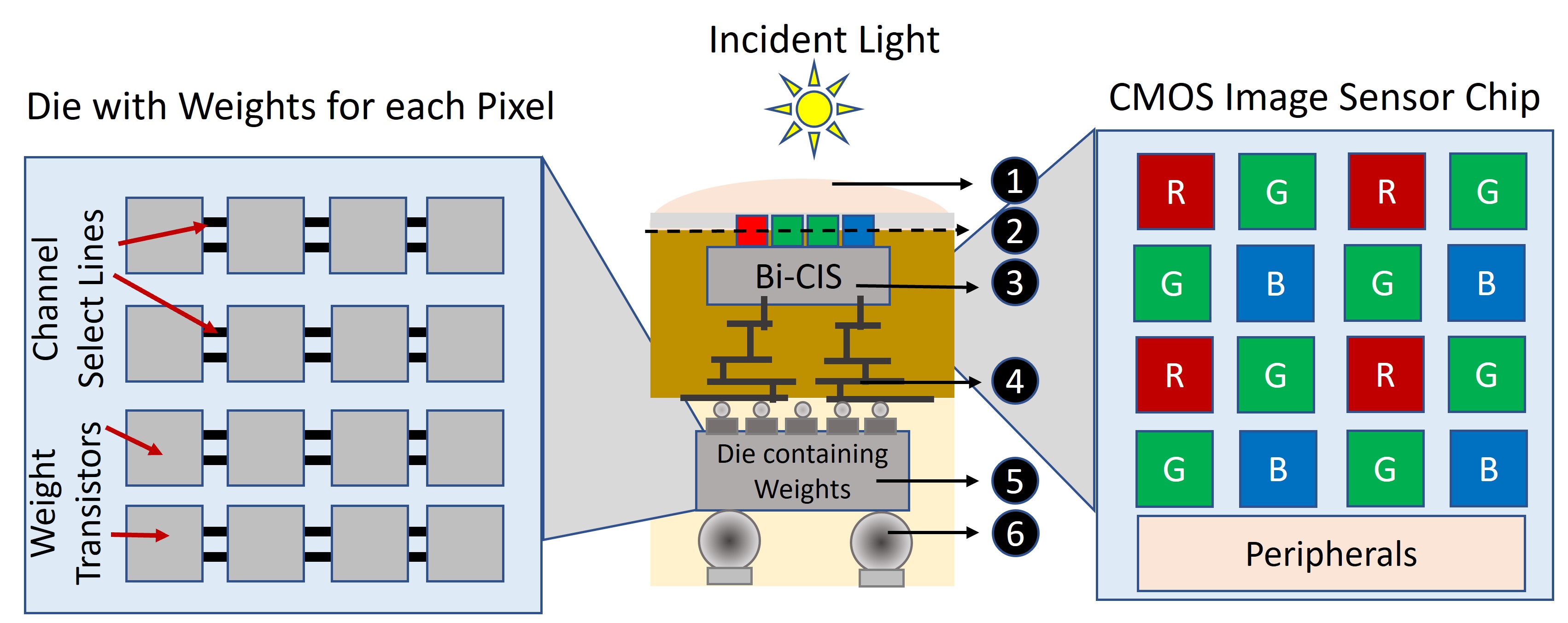}
\caption{ representative illustration of heterogeneously integrated system featuring P$^2$M paradigm, built on backside illuminated CMOS image sensor (Bi-CIS). \textcircled{\small{1}} Micro lens, \textcircled{\small{2}} Light shield, \textcircled{\small{3}} Backside illuminated CMOS Image Sensor (Bi-CIS), \textcircled{\small{4}} Backend of line of the Bi-CIS, \textcircled{\small{5}} Die consisting of weight transistors, \textcircled{\small{6}} solder bumps for input/output bus (I/O).}
\label{fig:p2m_fab}
\vspace{-1mm}
\end{figure}

In this section, we would like to highlight the viability of the proposed P$^2$M paradigm featuring memory-embedded pixels with respect to its manufacturability using existing foundry processes. A representative illustration of a heterogeneously integrated system catering to the needs of the proposed P$^2$M paradigm is shown in Fig. \ref{fig:p2m_fab}.
The figure consists of two key elements, i) backside illuminated CMOS image sensor (Bi-CIS), consisting of photo-diodes, read-out circuits and pixel transistors (reset, source follower and select transistors), and ii) a die consisting of multiple weight transistors per pixel (refer Fig \ref{fig:pip_circuit}). From Fig.  \ref{fig:pip_circuit}, it can be seen that each pixel consists of multiple weight transistors that would lead to exceptionally high area overhead. However, with the presented heterogeneous integration scheme of Fig. \ref{fig:p2m_fab}, the weight transistors are vertically aligned  below a standard pixel, thereby incurring no (or minimal) increase in footprint. Specifically, each Bi-CIS chip can be implemented in a leading or lagging technology node. The die consisting of weight transistors can be built on an advanced planar or non-planar technology node such that the multiple weight transistors can be accommodated in the same footprint occupied by a single pixel (assuming pixel sizes are larger than the weight transistor embedded memory circuit configuration). The Bi-CIS image sensor chip/die is heterogeneously integrated through a bonding process (die-to-die or die-to-wafer) 
integrating it onto the die consisting of weight transistors. Preferably, a die-to-wafer low-temperature metal-to-metal fusion with a dielectric-to-dielectric direct bonding hybrid process can achieve high-throughput sub-micron pitch scaling with precise vertical alignment \cite{gao2019} . One of the advantages of adapting this heterogeneous integration technology is that chips of different sizes can be fabricated at distinct foundry sources, technology nodes, and functions and then integrated together. In case there are any limitations due to the increased number of transistors in the die consisting of the weights, a conventional pixel-level integration scheme, such as Stacked Pixel Level Connections (SPLC), which shields the logic CMOS layer from the incident light through the Bi-CIS chip region, would also provide a high pixel density and a large dynamic range \cite{venezia2018}.
Alternatively, one could also adopt the \textit{through silicon via} (TSV) integration technique for front-side illuminated CMOS image sensor (Fi-CIS), wherein the CMOS image sensor is bonded onto the die consisting of memory elements through a TSV process. However, in the Bi-CIS, the wiring is moved away from the illuminated light path allowing more light to reach the sensor, giving better low-light performance \cite{6487825}.

Advantageously, the heterogeneous integration scheme can be used to manufacture P$^2$M sensor systems on existing as well as emerging technologies. Specifically, the die consisting of weight transistors could use a ROM-based structure as shown in Section \ref{sec:circuit_architecture} or other emerging programmable non-volatile memory technologies like PCM \cite{lee2010micro}, RRAM \cite{guo2018rram}, MRAM \cite{chih2020ISSCC}, ferroelectric field effect transistors (FeFETs) \cite{fefet} etc., manufactured in distinct foundries and subsequently heterogeneously integrated with the CIS die. 
Thus, the proposed heterogeneous integration allows us to achieve lower area-overhead, while simultaneously enabling seamless, massively parallel convolution.
As a result, individual pixels have in-situ access to both activation and weights as needed by the P$^2$M paradigm which obviates the need to transfer weights or activation from one physical location to another through a bandwidth constrained bus. Hence, unlike other multi-chip solutions \cite{sony2020vision}, our approach does not incur energy bottlenecks.



\begin{figure}[!t]
\centering
\includegraphics[width = 0.8\linewidth]{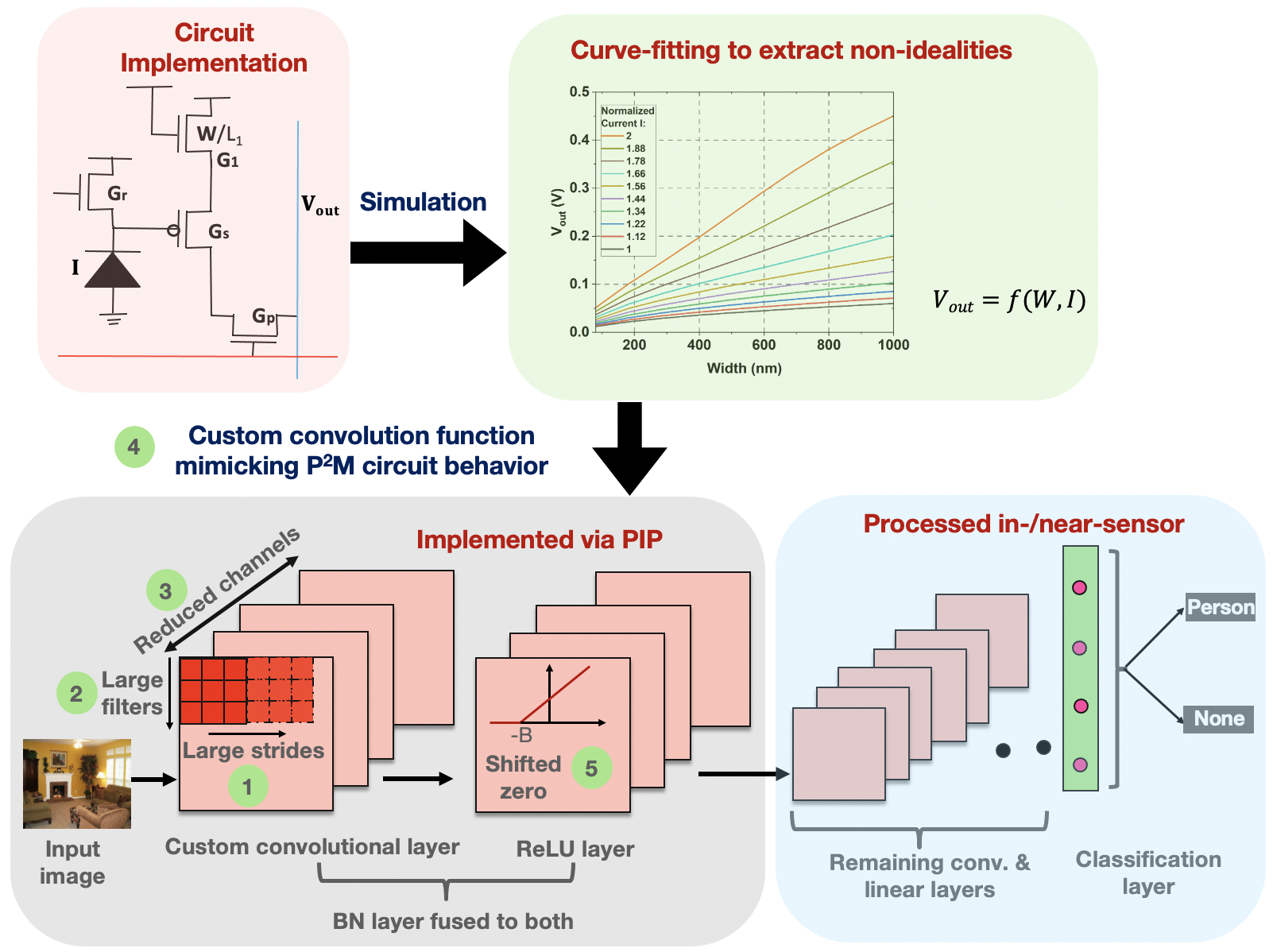}
\caption{Algorithm-circuit  co-design  framework  to  enable our proposed P$^2$M  approach optimize  both  the  performance  and energy-efficiency of vision workloads. We propose the use of \textcircled{\small{1}} large strides, \textcircled{\small{2}} large kernel sizes, \textcircled{\small{3}}reduced number of channels, \textcircled{\small{4}} P$^2$M custom convolution, and \textcircled{\small{5}} shifted ReLU operation to incorporate the shift term of the batch normalization layer, for emulating accurate P$^2$M circuit behaviour.}
\label{fig:pip_framework}
\vspace{-1mm}
\end{figure}

\section{P$^2$M-constrained Algorithm-Circuit Co-Design}\label{sec:algo_HW_codesign}

In this section, we present our algorithmic optimizations to standard CNN backbones that are guided by 1) P$^2$M circuit constraints arising due to analog computing nature of the proposed pixel array and the limited conversion precision of on-chip SS-ADCs, 2) the need for achieving state-of-the-art test accuracy, and 3) maximizing desired hardware metrics of high bandwidth reduction, energy-efficiency and low-latency of P$^2$M computing, and meeting the memory and compute budget of the VWW application. The reported improvement in hardware metrics (illustrated in Section \ref{subsec:edp}), is thus a result of intricate circuit-algorithm co-optimization.



\subsection{Custom Convolution for the First Layer Modeling Circuit Non-Idealities}

From an algorithmic perspective, the first layer of a CNN is a linear convolution operation followed by BN, and non-linear (ReLU) activation. The P$^2$M circuit scheme, explained in Section \ref{sec:circuit_architecture}, implements convolution operation in analog domain using modified memory-embedded pixels. The constituent entities of these pixels are transistors, which are inherently non-linear devices. As such, in general, any analog convolution circuit consisting of transistor devices will exhibit non-ideal non-linear behavior with respect to the convolution operation. 
Many existing works, specifically in the domain of memristive analog dot product operation, ignore non-idealities arising from non-linear transistor devices \cite{rxnn,memtorch}. In contrast, to capture these non-linearities, we performed extensive simulations of the presented P$^2$M circuit spanning wide range of circuit parameters like the width of weight transistors, photodiode current \textit{etc.} based on commercial $22$nm Globafoundries transistor technology node. The resulting SPICE results, \textit{i.e.} the pixel output voltages corresponding to a range of weights and photodiode currents, were modeled using a behavioral curve-fitting function. 
The generated function was then included in our algorithmic framework, replacing the convolution operation in the first layer of the network. 
In particular, we accumulate the output of the curve-fitting function, one for each pixel in the receptive field (we have $3$ input channels, and a kernel size of $5{\times}5$, and hence, our receptive field size is 75), to model each inner-product generated by the in-pixel convolutional layer. 
This algorithmic framework was then used to optimize the CNN training for the VWW dataset.

\subsection{Circuit-Algorithm Co-optimization of CNN Backbone subject to P$^2$M Constrains}\label{subsec:modified_cnn}

As explained in Section \ref{subsec:embedded_memory}, the P$^2$M circuit scheme maximizes parallelism and data bandwidth reduction by activating multiple pixels and reading multiple parallel analog convolution operations for a given channel in the output feature map. The analog convolution operation is repeated for each channel in the output feature map serially. Thus, parallel convolution in the circuit tends to improve parallelism, bandwidth reduction, energy-efficiency and speed. But, increasing the number of channels in the first layer increases the serial aspect of the convolution and degrades parallelism, bandwidth reduction, energy-efficiency, and speed. This creates an intricate circuit-algorithm trade-off, wherein the backbone CNN has to be optimized for having larger kernel sizes (that increases the concurrent activation of more pixels, helping parallelism) and non-overlapping strides (to reduce the dimensionality in the downstream CNN layers, thereby reducing the number of multiply-and-adds and peak memory usage), smaller number of channels (to reduce serial operation for each channel), while maintaining close to state-of-the-art classification accuracy and taking into account the non-idealities associated with analog convolution operation. Also, decreasing number of channels decreases the number of weight transistors embedded within each pixel (each pixel has weight transistors equal to the number of channels in the output feature map), improving area and power consumption. Furthermore, the resulting smaller output activation map (due to reduced number of channels, and larger kernel sizes with non-overlapping strides) reduces the energy incurred in transmission of data from the CIS to the downstream CNN processing unit and the number of floating point operations (and consequently, energy consumption) in downstream layers.


In addition, we propose to fuse the BN layer, partly in the preceding convolutional layer, and partly in the succeeding ReLU layer to enable its implementation via P$^2$M. Let us consider a BN layer with $\gamma$ and $\beta$ as the trainable parameters, which remain fixed during inference. During the training phase, the BN layer normalizes feature maps with a running
mean $\mu$ and a running variance $\sigma$. However, during inference, $\mu$ and $\sigma$ are computed from the mini-batch statistics and kept fixed \cite{NEURIPS2018_batchnorm}, and hence, the BN layer implements a linear function, as shown below.
\begin{equation}
Y=\gamma\frac{X-\mu}{\sqrt{\sigma^2+\epsilon}}+\beta=\left(\frac{\gamma}{\sqrt{\sigma^2+\epsilon}}\right)\cdot X+\left(\beta-\frac{\gamma\mu}{\sqrt{\sigma^2+\epsilon}}\right)=A\cdot X + B \label{eq:batch_norm}
\end{equation}
\noindent
We propose to fuse the scale term $A$ into the weights (value of the pixel embedded weight tensor is $A\cdot\theta$, where $\theta$ is the final weight tensor obtained by our training) that are embedded as the transistor widths in the pixel array. Additionally, we propose to use a shifted ReLU activation function, following the covolutional layer, as shown in Fig. \ref{fig:pip_framework} to incorporate the shift term $B$. We use the counter-based ADC implementation illustrated in Section \ref{subsec:relu_circuit} to implement the shifted ReLU activation. This can be easily achieved by resetting the counter to a non-zero value corresponding the the term $B$ at the start of the convolution operation, as opposed to resetting the counter to zero.

Moreover, to minimize the energy cost of the analog-to-digital conversion in our P$^2$M approach, we must also quantize the layer output to as few bits as possible subject to achieving the desired accuracy. 
We train a floating-point model with close to state-of-the-accuracy, and then perform quantization in the first convolutional layer to obtain low-precision weights and activations during inference \cite{NEURIPS2020_ebd9629f}. We also quantize the mean, variance, and the trainable parameters of the BN layer, as all these affect the shift term $B$ (please see Eq. \ref{eq:batch_norm}), that should be quantized for the low-precision shifted ADC implementation. We avoid  quantization-aware training \cite{courbariaux2016binarized} because it significantly increases the training cost with negligible reduction in bit-precision for our model. With the bandwidth reduction obtained by all these approaches, the output feature map of the P$^2$M-implemented layers can more easily be implemented in micro-controllers with extremely low memory footprint, while P$^2$M itself greatly improves the energy-efficiency of the first layer. Our approach can thus enable TinyML applications that usually have a tight compute and memory budget, as illustrated in Section \ref{subsec:benchmark}

\subsection{Quantification of bandwidth reduction}

To quantify the bandwidth reduction (BR) after the first layer obtained by P$^2$M (BN and ReLU layers do not yield any BR), let the number of elements in the RGB input image be $I$ and in the output activation map after the ReLU activation layer be $O$. Then, $BR$ can be estimated as
\begin{equation}
BR=\left(\frac{O}{I}\right)\left(\frac{4}{3}\right)\left(\frac{12}{N_b}\right)\label{eq:DR_1}
\end{equation}
\noindent
Here, the factor $\left(\frac{4}{3}\right)$ represents the compression from Bayer’s pattern of RGGB pixels to RGB pixels because we can either ignore the additional green pixel or design the circuit to effectively take the average of the photo-diode currents coming from the green pixels. The factor $\frac{12}{N_b}$ represents the ratio of the bit-precision between the image pixels captured by the sensor (pixels typically have a bit-depth of $12$ \cite{onsemi:AR0135AT}) and the quantized output of our convolutional layer denoted as $N_b$. Let us now substitute
\begin{equation}
O={\left(\frac{i-k+2*p}{s}+1\right)^2*c_{o}}, \quad I=i^2*3
\end{equation}
\noindent
into Eq. \ref{eq:DR_1}, where $i$ denotes the spatial dimension of the input image, and $k$, $p$, $s$ denote the kernel size, padding and stride of the in-pixel convolutional layer, respectively. These hyperparameters, along with $N_b$ are obtained via a thorough algorithmic design space exploration with the goal of achieving the best accuracy, subject to meeting the hardware constraints and the memory and compute budget of our TinyML benchmark. We show their values in Table \ref{tab:notations}, and substitute them in Eq. \ref{eq:DR_1} to obtain a BR of $21\times$. 

\renewcommand{\arraystretch}{1.4}
\begin{table}
\scriptsize\addtolength{\tabcolsep}{-4pt}
\begin{center}
\begin{tabular}{|c|c|}
\hline
Hyperparameter & Value\\
\hline \hline
kernel size of the convolutional layer ($k$) & 5  \\
\hline
padding of the convolutional layer ($p$) & 0 \\
\hline
stride of the convolutional layer ($s$) & 5 \\
\hline
number of output channels of the convolutional layer ($c_o$) & 8 \\
\hline
bit-precision of the P$^2$M-enabled convolutional layer output ($N_b$) & 8 \\
\hline
\end{tabular}
\end{center}
\caption{Model hyperparameters and their values to enable bandwidth reduction in the in-pixel layer.}
\label{tab:notations}
\end{table}

\section{Experimental Results}\label{sec:results}

\subsection{Benchmarking Dataset \& Model}\label{subsec:benchmark}

This paper focuses on the potential of P$^2$M for TinyML applications, \textit{i.e.}, with models that can be deployed on low-power IoT devices with only a few kilobytes of on-chip memory \cite{RAY2021,tinyml2,banbury2021micronets}. 
In particular, the Visual Wake Words (VWW) dataset \cite{chowdhery2019visual} presents a relevant use case for visual TinyML. It consists of high resolution images that include visual cues to “wake-up" AI-powered home assistant devices, such as Amazon's Astro \cite{astro2021}, that requires real-time inference in resource-constrained settings. The goal of the VWW challenge is to detect the presence of a human in the frame with very little resources -  close to $250$KB peak
RAM usage and model size \cite{chowdhery2019visual}. 
To meet these constraints, current solutions involve downsampling the input image to medium resolution ($224{\times}224$) which costs some accuracy \cite{NEURIPS2020_ebd9629f}.
We choose MobileNetV$2$ \cite{howard2017mobilenets} as our baseline CNN architecture with $32$ and $320$ channels for the first and last convolutional layers respectively that supports full resolution ($560{\times}560$) images. In order to avoid overfitting to only two classes in the VWW dataset, we decrease the number of channels in the last depthwise separable convolutional block by $3\times$.
MobileNetV2, similar to other MobileNet class of models, is very compact \cite{howard2017mobilenets} with size less than the maximum allowed in the VWW challenge.
It performs well on complex datasets like ImageNet \cite{russakovsky2015imagenet} and, as shown in Section \ref{sec:results}, does very well on VWWs.

To evaluate P$^2$M on MobileNetV2, we create a custom model that replaces the first convolutional layer with our P$^2$M custom layer that captures the systematic non-idealities of the analog circuits, the reduced number of output channels, and limitation of non-overlapping strides, as discussed in Section \ref{sec:algo_HW_codesign}. 

We train both the baseline and P$^2$M custom models in PyTorch using the SGD optimizer with momentum equal to $0.9$ for $100$ epochs. The baseline model has an initial learning rate (LR) of $0.03$, while the custom counterpart has an initial LR of $0.003$. Both the learning rates decay by a factor of $0.2$ at every $35$ and $45$ epochs. After training a floating-point model with the best validation accuracy, we perform quantization to obtain $8$-bit integer weights, activations, and the parameters (including the mean and variance) of the BN layer. All experiments are performed on a Nvidia $2080$Ti GPU with $11$ GB memory.

\subsection{Classification Accuracy}\label{subsec:classification_accuracy}

\textit{Comparison between baseline and P$^2$M custom models}: We evaluated the performance of the baseline and P$^2$M custom MobileNet-V2 models 
on the VWW dataset in Table \ref{tab:custom_baseline_acc}. Our baseline model currently yields the best test accuracy on the VWW dataset among the models available in literature that does not leverage any additional pre-training or augmentation. Note that our baseline model requires a significant amount of peak memory and MAdds (${\sim}30\times$ more than that allowed in the VWW challenge), however, serves a good benchmark for comparing accuracy. We observe that the P$^2$M-enabled custom model can reduce the number of MAdds by ${\sim}7.15\times$, and peak memory usage by ${\sim}25.1\times$ with $1.47\%$ drop in the test accuracy compared to the uncompressed baseline model for an image resolution of $560{\times}560$. With the memory reduction, our P$^2$M model can run on tiny micro-controllers with only $270$ KB of on-chip SRAM. 
Note that peak
memory usage is calculated using the same convention as \cite{chowdhery2019visual}. Notice also that both the baseline and custom model accuracies drop (albeit the drop is significantly higher for the custom model) as we reduce the image resolution, which highlights the need for high-resolution images and the efficacy of P$^2$M in both alleviating the bandwidth bottleneck between sensing and processing, and reducing the number of MAdds for the downstream CNN processing.

\renewcommand{\arraystretch}{1.4}
\begin{table}[!t]
\begin{center}
\scriptsize\addtolength{\tabcolsep}{-1pt}
\begin{tabular}{|c|c|c|c|c|}
\hline
  Image Resolution & Model & Test Accuracy ($\%$) & Number of MAdds (G) & Peak memory usage (MB)  \\ 
\hline
\hline
  560$\times$ 560  & baseline & 91.37 & 1.93  & 7.53  \\
        & P$^2$M custom & 89.90 & 0.27 & 0.30 \\
\hline
 225$\times$ 225 & baseline  & 90.56  &  0.31 & 1.2  \\
        & P$^2$M custom & 84.30  & 0.05 & 0.049   \\
\hline
 115$\times$ 115 & baseline & 91.10 & 0.09 & 0.311  \\
        & P$^2$M custom & 80.00 & 0.01 & 0.013  \\
\hline

\end{tabular}
\end{center}
\caption{Test accuracies, number of MAdds, and peak memory usage of baseline and P$^2$M custom compressed model while classifying on the VWW dataset for different input image resolutions.}
\label{tab:custom_baseline_acc}
\end{table}

\textit{Comparison with SOTA models}: Table \ref{tab:vww_sota_comparison}  provides a comparison of the performances of models generated through our algorithm-circuit co-simulation framework with SOTA TinyML models for VWW. Our P$^2$M custom models yield test accuracies within $0.37\%$ of the best performing model in the literature \cite{proxylessnas}. Note that we have trained our models solely based on the training data provided, whereas ProxylessNAS \cite{proxylessnas}, that won the 2019 VWW challenge leveraged additional pretraining with ImageNet. Hence, for consistency, we report the test accuracy of ProxylessNAS with identical training configurations on the final network provided by the authors, similar to \cite{NEURIPS2020_ebd9629f}. Note that \cite{zhou2021analognets} leveraged massively parallel energy-efficient analog in-memory computing to implement MobileNet-V2 for VWW, but incurs an accuracy drop of $5.67\%$ and $4.43\%$ compared to our baseline and the previous state-of-the-art \cite{proxylessnas} models. This probably implies the need for intricate algorithm-hardware co-design and accurately modeling of the hardware non-idealities in the algorithmic framework, as shown in our work.

\renewcommand{\arraystretch}{1.4}
\begin{table}
\begin{center}
\scriptsize\addtolength{\tabcolsep}{-1pt}
\begin{tabular}{|c|c|c|c|}
\hline
Authors & Description & Model architecture & Test Accuracy ($\%$) \\
\hline
\hline
Saha et al. (2020) \cite{NEURIPS2020_ebd9629f} & RNNPooling & MobileNetV2 & 89.65 \\
\hline
Han et al. (2019) \cite{proxylessnas} & ProxylessNAS & Non-standard architecture & 90.27  \\
\hline 
Banbury et al. (2021) \cite{banbury2021micronets} & Differentiable NAS & MobileNet-V2 & 88.75 \\
\hline
Zhoue et al. (2021) \cite{zhou2021analognets} & Analog compute-in-memory & MobileNet-V2 & 85.7 \\
\hline
This work & P$^2$M & MobileNet-V2 & 89.90 \\ 
\hline

\end{tabular}
\end{center}
\caption{Performance comparison of the proposed P$^2$M-compatible models with state-of-the-art deep CNNs on VWW dataset.}
\label{tab:vww_sota_comparison}
\end{table}

\begin{figure}
\centering
\begin{subfigure}{0.44\textwidth}
    \includegraphics[width = \textwidth]{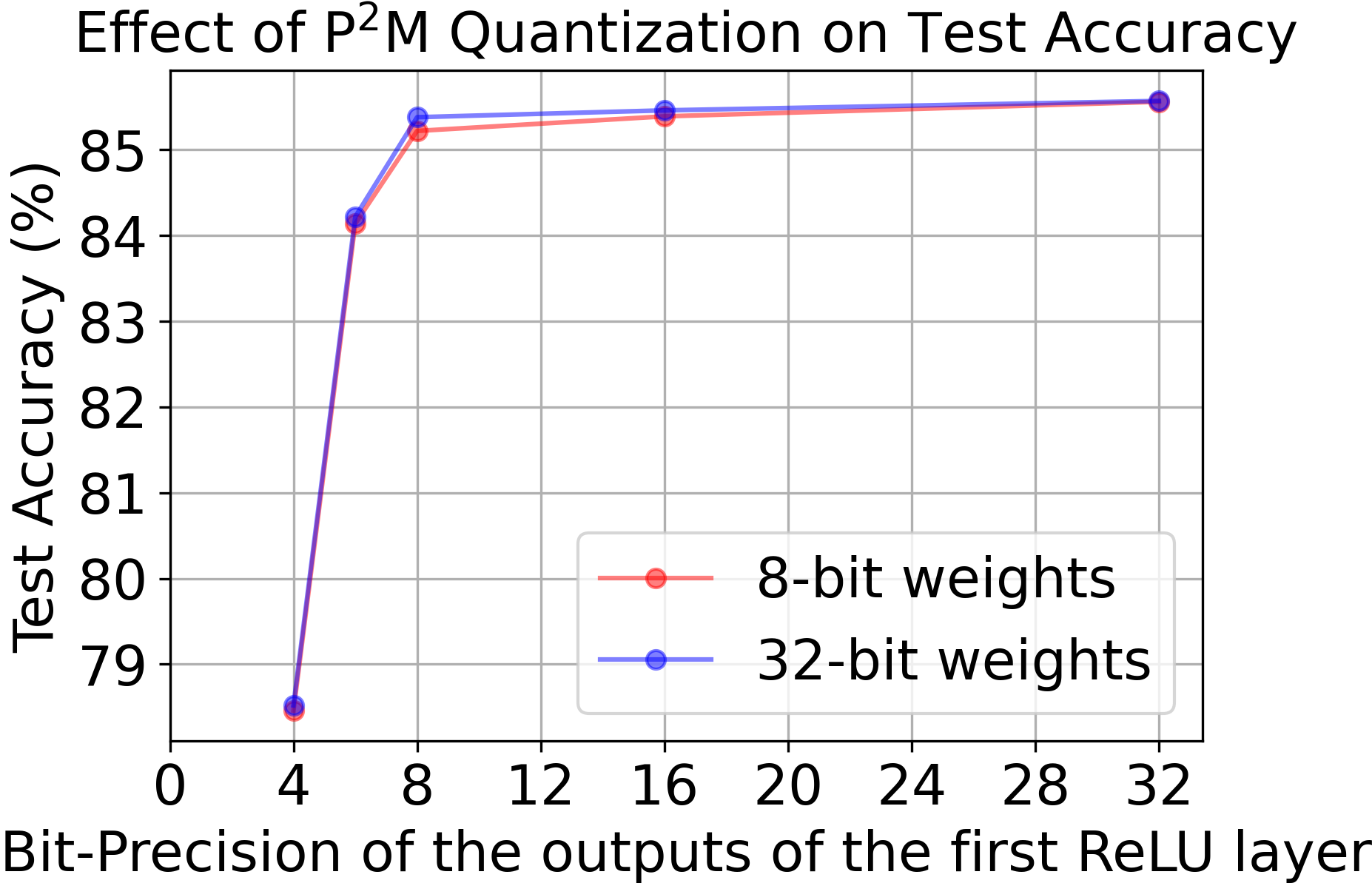}
    \caption{ }
    \label{fig:p2m_WI}
    
\end{subfigure}
\hfill
\begin{subfigure}{0.4\textwidth}
    \includegraphics[width = \textwidth]{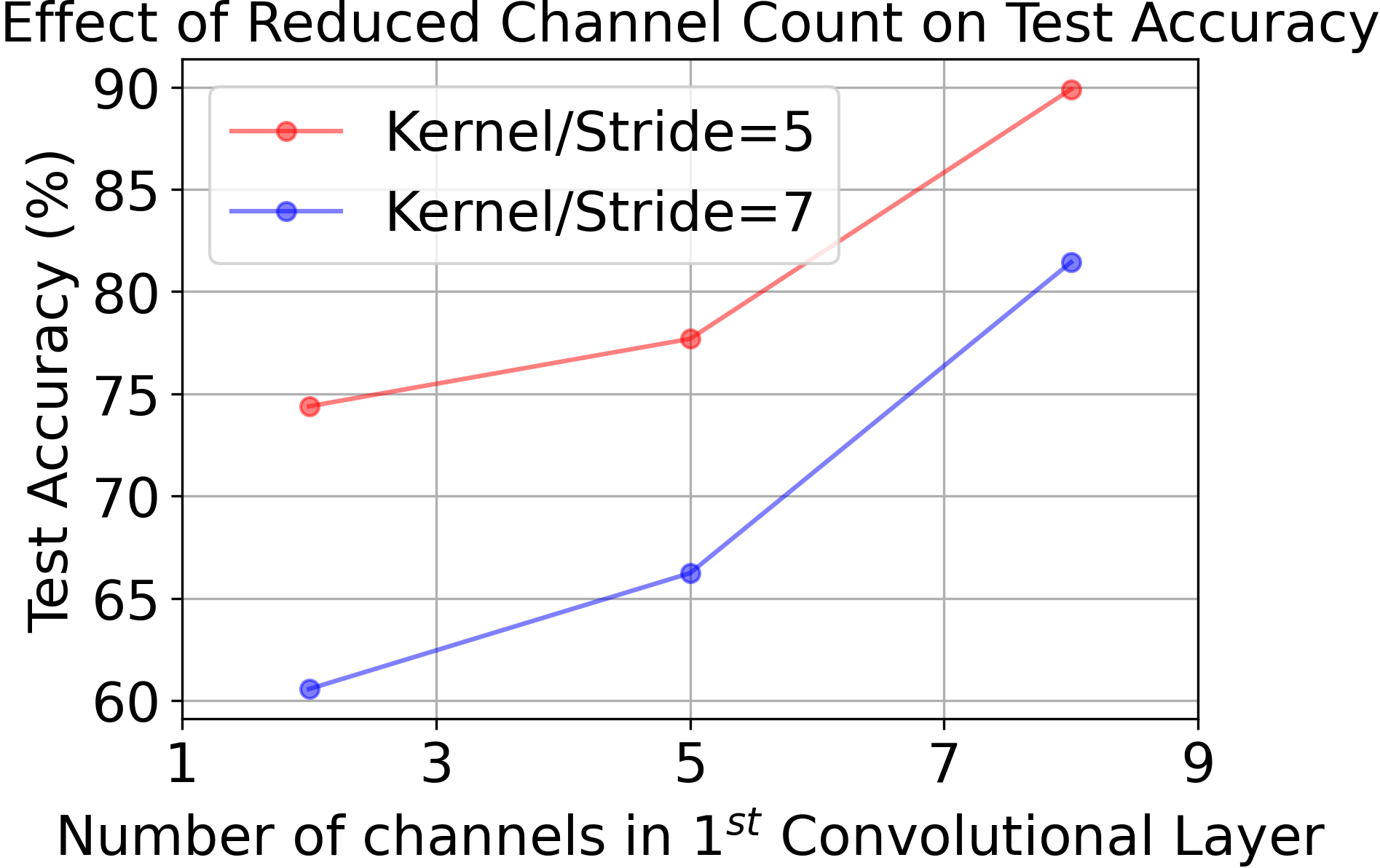}
    \caption{ }
    \label{fig:p2m_WI}
\end{subfigure}
\hfill
\caption{(a) Effect of quantization of the in-pixel output activations, and (b) Effect of the number of channels in the $1^{st}$ convolutional layer for different kernel sizes and strides, on the test accuracy of our P$^2$M custom model.}
\label{fig:quantization_vs_accuracy}
\vspace{-1mm}
\end{figure}

\textit{Effect of quantization of the in-pixel layer}: As discussed in Section \ref{sec:algo_HW_codesign}, we quantize the output of the first convolutional layer of our proposed model after training to reduce the power consumption due to the sensor ADCs and compress the output as outlined in Eq. \ref{eq:DR_1}. We sweep across output bit-precisions of \{$4$,$6$,$8$,$16$,$32$\} to explore the trade-off between accuracy and compression/efficiency as shown in Fig. \ref{fig:quantization_vs_accuracy}(a). We choose a bit-width of $8$ as it is the lowest precision that does not yield any accuracy drop compared to the full-precision models. As shown in Fig. \ref{fig:quantization_vs_accuracy}, the weights in the in-pixel layer can also be quantized to $8$ bits with an $8$-bit output activation map, with less than $0.1\%$ drop in accuracy.

\textit{Ablation study}: We also study the accuracy drop incurred due to each of the three modifications (non-overlapping strides, reduced channels, and custom function) in the P$^2$M-enabled custom model. Incorporation of the non-overlapping strides (stride of 5 for $5{\times}5$ kernels from a stride of 2 for $3{\times}3$ in the baseline model) leads to an accuracy drop of $0.58\%$. Reducing the number of output channels of the in-pixel convolution by $4{\times}$ ($8$ channels from $32$ channels in the baseline model), on the top of non-overlapping striding, reduces the test accuracy by $0.33\%$. Additionally, replacing the element-wise multiplication with the custom P$^2$M function in the convolution operation reduces the test accuracy by a total of $0.56\%$ compared to the baseline model. Note that we can further compress the in-pixel output by either increasing the stride value (changing the kernel size proportionately for non-overlapping strides) or decreasing the number of channels. But both of these approaches reduce the VWW test accuracy significantly, as shown in Fig. \ref{fig:quantization_vs_accuracy}(b). 

\begin{table}[!h]
\begin{center}
\scriptsize\addtolength{\tabcolsep}{-0pt}
\begin{tabular}{|c|c|c|c|c|c|}
\hline
 Model type & Sensing (pJ)  & ADC (pJ) & SoC comm. (pJ)   & MAdds (pJ) & Sensor output  \\
 {} & {($e_{pix}$)}  & ($e_{adc}$) & ($e_{com}$)  & ($e_{mac}$) & pixel ($N_{pix}$) \\ 
\hline
\hline
 P$^2$M (ours) & 148  & 41.9 & \multirow{3}{*}{900}  &  \multirow{3}{*}{1.568} & $112{\times}112{\times}8$ \\
\cline{1-3} \cline{6-6}
 Baseline (C) & \multirow{2}{*}{312}  & \multirow{2}{*}{86.14}  &   &  & \multirow{2}{*}{$560{\times}560{\times}3$}\\
 \cline{1-1} 
 Baseline (NC) & {}  & {}  &   &  & \\
 \hline
\end{tabular}
\end{center}
\caption{Energy estimates for different hardware components. The energy values are measured for designs in $22$nm CMOS technology. For the $e_{mac}$, we convert the corresponding value in $45$nm to that of $22$nm by following standard scaling strategy \cite{stillmaker2017scaling}.}
\label{tab:energy_and_model_estimates}
\vspace{-2mm}
\end{table}

\begin{table}[!h]
\begin{center}
\scriptsize\addtolength{\tabcolsep}{-0pt}
\begin{tabular}{|c|c|c|}
\hline
 Notation & Description & Value \\ 
\hline
\hline
 $B_{IO}$ & I/O band-width & 64 \\
 \hline
 $B_{W}$ & Weight representation bit-width & 32 \\
 \hline
 $N_{bank}$ & Number of memory banks & 4 \\
 \hline
 $N_{mult}$ & Number of multiplication units & 175 \\
  \hline
 $T_{sens}$ & Sensor read delay & 35.84 ms (P$^2$M)\\
 {} & {} & 39.2 ms (baseline)\\
   \hline
 $T_{adc}$ & ADC operation delay & 0.229 ms (P$^2$M)\\
 {} & {} & 4.58 ms (baseline)\\
  \hline
 $t_{mult}$ & Time required to perform 1 mult. in SoC & 5.48 ns \\
   \hline
 $t_{read}$ & Time required to perform 1 read from SRAM in SoC & 5.48 ns \\
 \hline
\end{tabular}
\end{center}
\caption{The description and values of the notations used for computation of delay. Note that we  strategy and calculated the delay in $22$nm technology for $32$-bit read and MAdd operations by applying standard technology scaling rules initial values in $65$nm technology \cite{ali2020imac}. We directly evaluated the $T_{read}$ and $T_{adc}$ through circuit simulations in $22$nm technology node.}
\label{tab:delay_variable_values}
\vspace{-0mm}
\end{table}

\subsection{EDP Estimation}\label{subsec:edp}

We develop a circuit-algorithm co-simulation framework to characterize
the energy and delay of our baseline and P$^2$M-implemented VWW models.
The total energy consumption for both these models can be partitioned into three major components: sensor ($E_{sens}$), sensor-to-SoC communication ($E_{com}$), and SoC energy ($E_{soc}$). Sensor energy can be further decomposed to pixel read-out ($E_{pix}$) and analog-to-digital conversion (ADC) cost ($E_{adc}$). $E_{soc}$, on the other hand, is primarily composed of the MAdd operations ($E_{mac}$) and parameter read ($E_{read}$) cost. Hence, the total energy can be approximated as:
\begin{align}
E_{tot} \approx \underbrace{(e_{pix}+e_{adc})*N_{pix}}_{E_{sens}} + \underbrace{e_{com}*N_{pix}}_{E_{com}} + \underbrace{e_{mac}*N_{mac}}_{E_{mac}} + \underbrace{e_{read}*N_{read}}_{E_{read}}.
\end{align}
\noindent
Here, $e_{sens}$ and $e_{com}$ represents per-pixel sensing and communication energy, respectively. $e_{mac}$ is the energy incurred in one MAC operation, $e_{read}$ represents a parameter's read energy, and $N_{pix}$ denotes the number of pixels communicated from sensor to SoC. For a convolutional layer that takes an input $\mathbf{I} \in R^{h_i{\times}w_i{\times}c_i}$ and weight tensor $\mathbf{\theta} \in R^{k{\times}k{\times}c_i{\times}c_o}$ to produce output $\mathbf{O} \in R^{h_o{\times}w_o{\times}c_o}$, the $N_{mac}$ \cite{kundu2020pre} and $N_{read}$ can be computed as,
\begin{equation}
    N_{mac} = h_o*w_o*k^2*c_i*c_o 
\end{equation}
\vspace{-7mm}
\begin{equation}
    N_{read} = k^2*c_i*c_o
\end{equation}

\begin{figure}[!t]
\centering
\includegraphics[width = 0.8\linewidth]{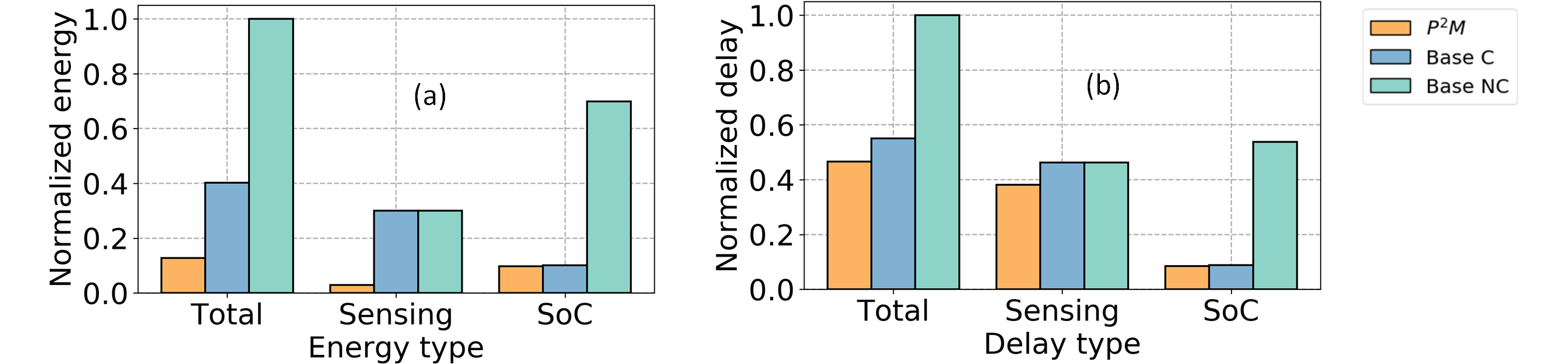}
\caption{Comparison of normalized \textit{total}, \textit{sensing}, and \textit{SoC} (a) energy cost and (b) delay between the P$^2$M, and baseline models architectures (compressed C, and non-compressed NC). Note, the normalization of each component was done by diving the corresponding  energy (delay) value with the maximum total energy (delay) value of the three components.}
\label{fig:pip_vs_base_energy}
\vspace{-1mm}
\end{figure}

The energy values we have used to evaluate $E_{tot}$ are presented in Table \ref{tab:energy_and_model_estimates}. While $e_{pix}$ is obtained from our circuit simulations, $e_{adc}$ and $e_{com}$ are obtained from \cite{gonugondla2021imc} and \cite{kodukula2020sensors} respectively. We ignore the value of $E_{read}$ as it corresponds to only a small fraction (${<}10^{-4}$) of the total energy, similar to \cite{Kundu_2021_WACV,datta2021ijcnn,datta2022date,kundu2021hire}. Fig. \ref{fig:pip_vs_base_energy}(a) shows the comparison of energy costs for standard vs P$^2$M-implemented models. In particular, P$^2$M can yield an energy reduction of up to $7.81\times$. Moreover, the energy savings is larger when the feature map needs to be transferred from an edge device to the cloud for further processing, due to the high communication costs. 
Note, here we assumed two baseline scenarios one with compression and one without compression. The first baseline is MobileNetV2 which aggressively down-samples the input similar to P$^2$M ($h_i/w_i: 560 \longrightarrow h_o/w_o: 112$). For the second baseline model,
we assumed standard first layer convolution kernels causing standard feature down-sampling ($h_i/w_i: 560 \longrightarrow h_o/w_o: 279$). 

To evaluate the delay of the models we assume sequential execution of the layer operations \cite{ali2020imac,kang2018memory,datta2021hsi} and compute a single convolutional layer delay as \cite{ali2020imac}
\begin{align}
    t_{conv} \approx \lceil \frac{(k)^2c_ic_o}{(B_{IO}/B_W)N_{bank}} \rceil*t_{read} + \lceil \frac{(k)^2c_ic_o}{N_{Mult}} \rceil h_o*w_o*t_{mult}.
    \label{eq:conv_dealay}
\end{align}
Based on this sequential assumption, the approximate compute delay for a single forward pass for our P$^2$M model can be given by
\begin{align}
    T_{delay} \approx T_{sens} + T_{adc} + T_{conv}.
\end{align}
Here, $T_{sens}$ and $T_{adc}$ correspond to the delay associated to the sensor read and ADC operation respectively. $T_{conv}$ corresponds to the delay associated with all the convolutional layers where each layer's delay is computed by Eq. \ref{eq:conv_dealay}. Fig. \ref{fig:pip_vs_base_energy}(b) shows the comparison of delay between P$^2$M and the corresponding baselines where the total delay is computed with the sequential sensing and SoC operation assumption. In particular, the proposed P$^2$M approach can yield an improved delay of up to $2.15\times$. Thus the total EDP advantage of P$^2$M can be up to $16.76\times$. On the other hand, even with the conservative assumption of total delay is estimated as \texttt{max}($T_{sens}$+$T_{adc}$, $T_{conv}$), the EDP advantage can be up to $\mathord{\sim}11\times$. 

\section{Conclusions}\label{sec:conc}
With the increased availability of high-resolution image sensors, there has been a growing demand for energy-efficient on-device AI solutions. To mitigate the large amount of data transmission between the sensor and the on-device AI accelerator/processor, we propose a novel paradigm called \textit{Processing-in-Pixel-in-Memory} (P$^2$M) which leverages advanced CMOS technologies to enable the pixel array to perform a wider range of complex operations, including many operations required by modern convolutional neural networks (CNN) pipelines, such as multi-channel, multi-bit convolution, BN and ReLU activation. Consequently, only the compressed meaningful data, for example after the first few layers of custom CNN processing, is transmitted downstream to the AI processor, significantly reducing the power consumption associated with the sensor ADC and required data transmission bandwidth. 
Our experimental results yield reduction of data rates after the sensor ADCs by up to ${\sim}21\times$ compared to standard near-sensor processing solutions, significantly reducing the complexity of downstream processing. This, in fact, enables the use of relatively low-cost micro-controllers for many low-power embedded vision applications and unlocks a wide range of visual TinyML applications that require high resolution images for accuracy, but are bounded by compute and memory usage. We can also leverage P$^2$M for even more complex applications, where downstream processing can be implemented using existing 
near-sensor computing techniques that leverage advanced $2.5$ and $3$D integration technologies \cite{Amir20183DSH}.


\section*{Acknowledgements}

We would like to acknowledge the DARPA HR$00112190120$ award for supporting this work. The views and conclusions contained herein are
those of the authors and should not be interpreted as necessarily representing the official policies or endorsements,
either expressed or implied, of DARPA.

\section*{Author contributions statement}

GD and SK proposed the use of P$^2$M for TinyML applications, developed the baseline and P$^2$M-constrained models, and analyzed their accuracies. GD and SK analyzed the EDP improvements over other standard implementations with the help of AJ$^{2}$ and ZY. AJ$^{1}$ and AJ$^{2}$ proposed the idea of P$^2$M and ZY and RL developed the corresponding circuit simulation framework. JM helped to incorporate the non-ideality in the P$^2$M layer in the ML framework. GD and AJ$^{2}$ wrote majority of the paper, while SK, AJ$^{1}$ and ZY wrote the remaining portions. AJ$^{1}$  helped in manufacturing feasibility analysis and proposed the use of heterogeneous integration scheme for P$^{2}$M. PB supervised the research and edited the manuscript extensively. All authors reviewed the manuscript. Note that AJ$^{1}$ and AJ$^{2}$ are Ajey P. Jacob and Akhilesh R. Jaiswal respectively.

\bibliography{sample}

\end{document}